\documentclass[letterpaper]{article} 

\usepackage{amsmath,amsfonts,bm}

\def\eqref#1{equation~\ref{#1}}

\def\1{\bm{1}}

\DeclareMathAlphabet{\mathsfit}{\encodingdefault}{\sfdefault}{m}{sl}
\SetMathAlphabet{\mathsfit}{bold}{\encodingdefault}{\sfdefault}{bx}{n}

\usepackage{amsmath,amsfonts,bm}

\newcommand{\grad}{\nabla}

\newcommand{\bI}{\mathbf{I}}

\newcommand{\bzero}{\mathbf{0}}

\newcommand{\bx}{\mathbf{x}}

\newcommand{\bz}{\mathbf{z}}

\newcommand{\bepsilon}{{\boldsymbol{\epsilon}}}
\newcommand{\brho}{{\boldsymbol{\rho}}}

\usepackage{aaai23}  
\usepackage{times}  
\usepackage{helvet}  
\usepackage{courier}  
\usepackage[hyphens]{url}  
\usepackage{graphicx} 
\usepackage{natbib}  
\usepackage{caption} 
\usepackage{booktabs}
\usepackage{scrextend}
\usepackage{mathtools}
\usepackage{graphicx}
\usepackage{subcaption}
\usepackage{enumitem}
\usepackage{colortbl}
\usepackage{wrapfig}
\usepackage{kotex}
\usepackage{lipsum}
\usepackage{xcolor,pifont}
\usepackage{algpseudocode,algorithm,algorithmicx,}
\usepackage{multirow}
\usepackage{newfloat}
\usepackage{listings}
\DeclareCaptionStyle{ruled}{labelfont=normalfont,labelsep=colon,strut=off} 
\floatstyle{ruled}
\newfloat{listing}{tb}{lst}{}
\floatname{listing}{Listing}
\newcommand{\stdv}[1]{\scriptsize$\pm$#1}

\title{VIDM: Video Implicit Diffusion Models}
\author{
    Kangfu Mei,
    Vishal M. Patel
}
\affiliations{
    Johns Hopkins University\\
    \url{https://kfmei.page/vidm/}
}

\usepackage{bibentry}

\begin{document}


\twocolumn[{%
\renewcommand\twocolumn[1][]{#1}%
\maketitle
\begin{center}
\vspace{-4.5\baselineskip}
\centerline{\includegraphics[width=.82\textwidth]{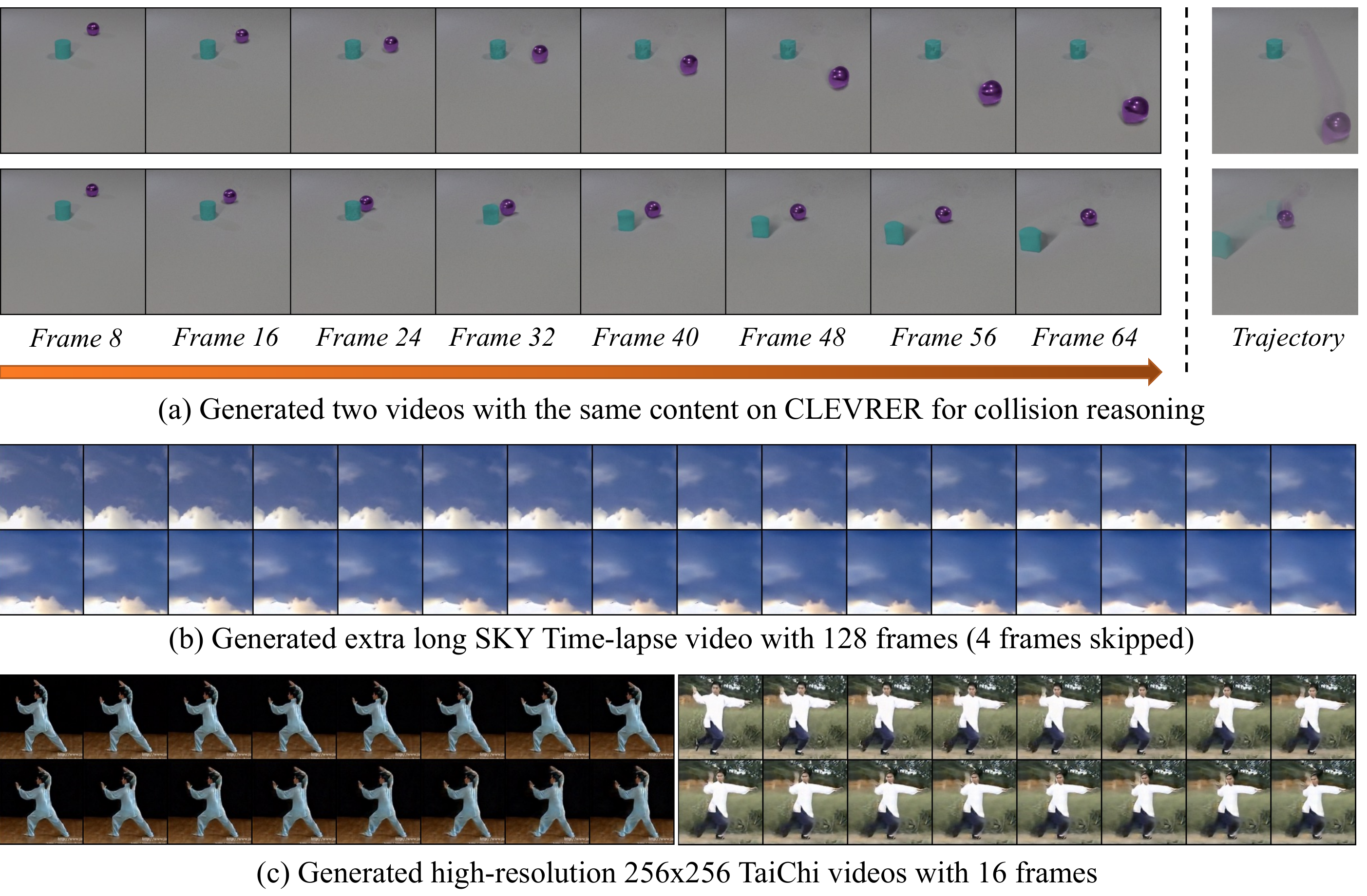}}
\vspace{-.5\baselineskip}
\captionof{figure}{Sample results corresponding to our method on multiple video datasets.}
\end{center}%
}]

\begin{abstract}
Diffusion models have emerged as a powerful generative method for synthesizing high-quality and diverse set of images.  In this paper, we propose a video generation method based on diffusion models, where the effects of motion are modeled in an implicit condition manner, i.e. one can sample plausible video motions according to the latent feature of frames.  We improve the quality of the generated videos by proposing multiple strategies such as sampling space truncation, robustness penalty, and positional group normalization.  Various experiments are conducted on datasets consisting of videos with different resolutions and different number of frames.  Results show that the proposed method outperforms the state-of-the-art generative adversarial network-based methods by a significant margin in terms of FVD scores as well as perceptible visual quality. 
\end{abstract}

\section{Introduction}
Image generation has gained significant traction since the introduction of Generative Adversarial Networks (GANs) \cite{goodfellow2014generative}.  In these methods, the idea is to generate new images that conform to the training data distribution.  Following the success of image synthesis, video generation has also gained significant attention. Various video generation methods have been proposed in the literature including GAN-based methods \cite{ vondrick2016generating, saito2017temporal, tulyakov2018mocogan, yu2022generating}, Autoregressive models~\cite{Weissenborn2020Scaling}, and Time-series models~\cite{tian2021good, skorokhodov2021stylegan}.  An advantage of some of these generative models is that they can learn to synthesize high-quality videos without requiring any labels.
These generative models have been shown to be beneficial in various high-level recognition tasks~\cite{srivastava2015unsupervised, vondrick2016generating}.

A GAN-based video generation model was proposed by \citet{vondrick2016generating}, which makes use of a spatio-temporal convolutional architecture and untangles the scene’s foreground from the background.
Another work proposed by \citet{tulyakov2018mocogan} is a continuous-time video generator. In this method, a video is decomposed into the content and motion vectors at generation and discriminated coherently by the discriminators.  While these GAN-based methods can model plausible moving objects and scenes, a better video generation model should be able to model the distribution of internal spatial and temporal changes with regards to the video content.

Different from GANs, diffusion models~\cite{sohl2015deep, ho2020denoising} as well as score-based models~\cite{song2020score} that model the probability directly have emerged as the new state-of-the-art generative models, and they have been shown to outperform GANs in various generation tasks \cite{dhariwal2021diffusion}. By learning to reverse the diffusion process that adds noise to data in finite successive steps, diffusion models can gradually map a Gaussian distribution to the probability distribution corresponding to a real complex high-dimensional dataset.
In its denoising process, conditional features like class labels of data can be applied to the network for specializing its sampling process.
By appropriately using conditional features, diffusion models have shown impressive performance in various applications, \emph{e.g.}, image debluring~\cite{whang2021deblurring} that conditions on the image residual, high-resolution image generation~\cite{ho2022cascaded} that conditions on the low-resolution images, and image editing~\cite{choi2021ilvr} that conditions on the style.
With more expressive conditional features like CLIP embeddings~\cite{radford2021learning}, diffusion models like DALLE-2~\cite{ramesh2022hierarchical} are capable of generating highly creative images with impressive photorealism.
But the condition mechanism in diffusion models is non-trivial and requires careful design to improve the quality of the generated images.

We assume that the subspace of a real video can be represented as a subspace of the video content, and the video motion is then generated by traversing point on the video content subspace.
Accurately modeling the content subspace increases the realism of frames, while accurately modeling the subspace of the trajectory regarding the video content can produce continuous and smooth video.  Thus a  better video generation model should own delicate modulation capability for simulating both the trajectory and realistic content.

Following this idea, we propose to model the video content and motion with two diffusion models separately.
The first video frame is generated by the content generator. Subsequently, the motion generator generates the next video frame based on the latent map of the first frame and the latest frame, i.e., an optical-flow like feature between the first and the latest frame estimated by an additional network.
This enables implicit modeling of dynamics by conditioning on the latent features.
After training, the optimized condition can best represent the spatial and temporal changes for generating the next frame.
By iteratively running the motion generator, the final video is generated in an autoregressive manner.
We experimentally find that the estimated condition significantly enhances the modeling capability of diffusion models,
Such an expressive model is capable of simulating the trajectory of videos according to the conditional latent.

The major idea of  our video implicit diffusion models is: 
\begin{itemize}
  \item \emph{Content Generator:} We propose to learn video content separately with an introduced diffusion model on video frames. It simplifies video generation modeling and provides easy scalability of complex models.
  Two heuristic mechanisms, including constant truncation and robustness penalty, are proposed for further improving its performance.  
  \item \emph{Motion Generator:} We propose a motion generator for modeling spatial and temporal changes. It can generate future frames according to the generated content in an autoregressive way.
  The generator is implicitly conditioned on the latent code predicted by a module similar to an optical-flow network.
  Furthermore, the coherency of spatial and temporal changes is regularized with an introduced positional group normalization, and the learning is simplified with our proposed adaptive feature residual.
\end{itemize}

The effectiveness of the proposed model is demonstrated on various datasets by comparing the performance with several state-of-the-art works, including very recent works MoCoGAN-HD~\cite{tian2021good}, DIGAN~\cite{yu2022generating}, and StyleGAN-V~\cite{skorokhodov2021stylegan}.  It is shown that our method achieves significantly better quantitative performance of Fr\'echet video distance and is experimentally observed to be capable of generating more realistic results.

\section{Denoising Diffusion Probabilistic Model}
Based on the success of diffusion-based models, we extend the generation process from 2D images into 3D videos and keep the modification as minimal as possible.
Our approach  is based on Denoising Diffusion Probabilistic Mdoel (DDPM) proposed by \citet{ho2020denoising} and its variant Guided-DDPM from \citet{dhariwal2021diffusion}.

\noindent \textbf{Learning Process.}
DDPM models the distribution of images $\bx_0\sim q(\bx_0)$ in a denoising process, and it learns noise $\bepsilon \sim \mathcal{N}(\mathbf{0}, \mathbf{I})$ with respect to timesteps $t$ (out of $T$) and defines noisy image $\bx_t$ as a function $\bepsilon_{\theta}(\bx_t, t)$, which is implemented as a modified U-Net~\cite{salimans2017pixelcnn++} $\bepsilon_\theta(\cdot)$ with parameters $\theta$.
A simplified learning objective is
\begin{align}
  \mathcal{L}_\theta(\epsilon, \bx_t, t) = \|\bepsilon_\theta(\bx_t, t) - \bepsilon\|.
\end{align}

\noindent \textbf{Noise Definition.}
Various attempts have been made to improve the form of $\bx_t$.
The basic formulation comes by combining the noise $\bepsilon \sim \mathcal{N}(\mathbf{0}, \mathbf{I})$ with a clean image $\bx_0$ in $t$ steps according to some pre-defined noise schedules $\alpha_t$ and its variant $\bar\alpha_t$ as $\bx_t = \sqrt{\bar\alpha_t}\bx_0 + \sqrt{1-\bar\alpha_t}\bepsilon$.

\begin{figure*}[htbp]
  \centering
  \includegraphics[width=.9\linewidth]{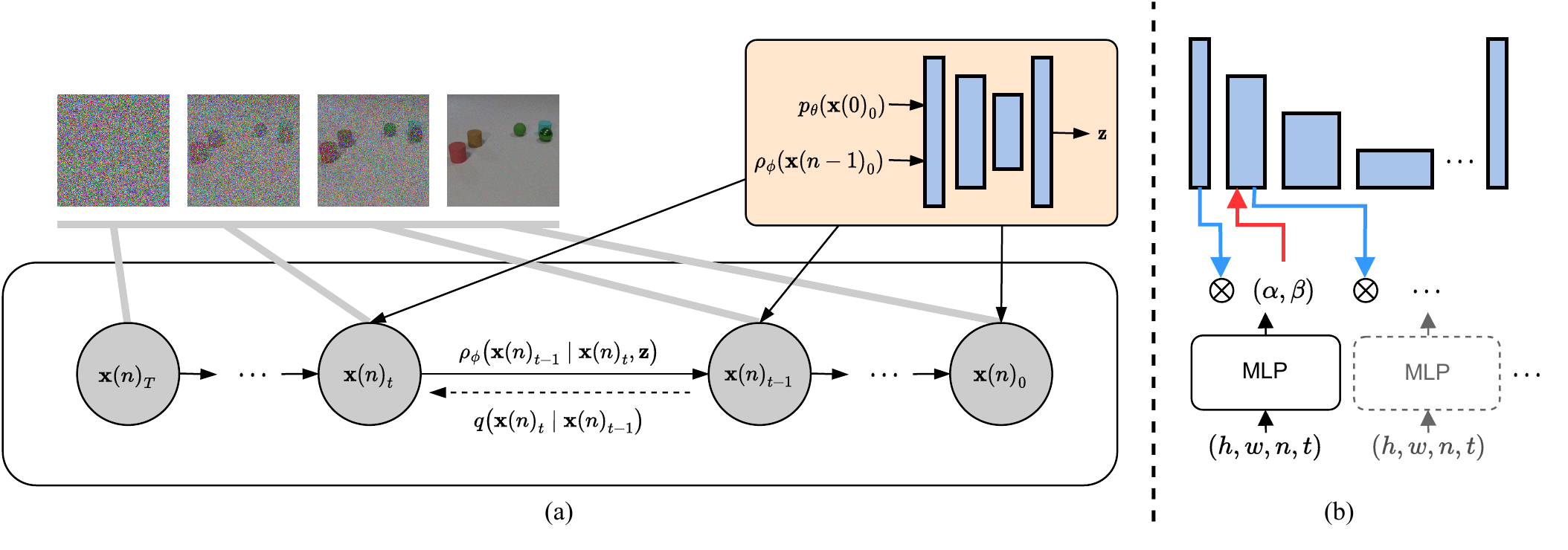}
  \vspace{-1\baselineskip}
  \caption{(a) Illustration of our graphical model at the $n$-th video frame sampling process. (b) The proposed positional group normalization concept when it is  applied to the diffusion network.}
  \vspace{-.5\baselineskip}
  \label{fig:pipeline}
\end{figure*}

\noindent \textbf{Inference Process.} The generation process starts from noise $\bx_T \sim \mathcal{N}(\mathbf{0}, \mathbf{I})$ with random noise $\bepsilon$ and predefined variance $\sigma_t$ and is defined as
\begin{align}
  \label{eq:inference}
  \bx_{t-1} = \frac{1}{\sqrt{\alpha_t}}\left(\bx_t - \frac{1-\alpha_t}{\sqrt{1-\bar\alpha_t}}\bepsilon_\theta(\bx_t, t)\right) + \sigma_t\bepsilon.
\end{align}

\noindent \textbf{Truncation Trick.}
Compared with the GAN-based methods, though the superiority in the diversity of generation is achieved, diffusion models often result in the generation of poor quality of objects.  
Inspired by the heavily explored approach used by GANs, that is sampling from a truncated or a shrunk sampling space~\cite{brock2018large, kingma2018glow, karras2019style}, we propose to truncate noise $\bepsilon$ implicitly.

Inspired by the practice of StyleGAN~\cite{karras2019style}, which starts from a learnable constant and then gradually upsamples the features until the final output layer, we propose to concatenate noisy image $\bx_t$ with a learnable constant $c$ that has the same dimension as $\bx_t$ at each diffusion step.
Such a strategy truncates the sampling space of the noise in an implicit way without modifying the network architecture, and the learning objective is slightly changed as $\|\bepsilon_\theta(\bx_t, c, t) - \bepsilon\|$.
During inference, the constant $c$ is fixed and the inference process~\eqref{eq:inference} is subsequently updated as
\begin{align}
  \bx_{t-1} = \frac{1}{\sqrt{\alpha_t}}\left(\bx_t - \frac{1-\alpha_t}{\sqrt{1-\bar\alpha_t}}\bepsilon_\theta(\bx_t, c, t)\right) + \sigma_t\bepsilon.
\end{align}

\noindent \textbf{Robustness Penalty.}\label{sec:rob}
Dropout layer has been empirically applied in DDPM  for suppressing overfitting artifacts.
However, the practice of applying dropout depends on the dataset  and harms the general performance in most cases.
We observed that  overfitting not only depends on the dataset but different classes of the same dataset as well and thus dropout is conventionally avoided.

To enable an adaptive strategy for preventing overfitting, we propose to add a penalty function~\cite{charbonnier1994two} at the learning objective instead of dropout layers as
\begin{align}
  \label{eq:newobj}
  \mathcal{L}_\theta(\bepsilon, \bx_t, t) = \sqrt{(\bepsilon_{\theta}(\bx_t, c, t) - \bepsilon)^2 + \eta^2},
\end{align}
where $\eta$ is a constant that is experimentally set as $1e-8$, while the other settings, including $\{1e-5, 1e-6, 1e-7\}$, haven't shown significantly better performance.
Such a modification doesn't hurt the differentiability of the original learning objective.

\section{Video Implicit Diffusion Model}
Our proposed video generation method consists of two streams for content and motion generation, respectively.
The two streams share a similar network architecture but different in learning objectives.  In addition, they have different conditions which helps to keep the design redundancy minimal and reduces the optimization cost. 
We denote the $n$-th frame of the $N$-frame video as $\bx(n)_0$. The noisy frame at the $t$th timestep is denoted as $\bx(n)_t$.

\noindent \textbf{Content Generator} models the distribution of random video frames $\bx_0\sim q(\bx_0)$ with a network $\bepsilon_\theta(\cdot)$ and is truncated by constant tensor $c$.
The frame $\bx_0$ is randomly selected from videos  without specification.
The network $\bepsilon_\theta(\cdot)$ is the modified U-Net proposed by \citet{dhariwal2021diffusion} with \emph{Multi-Head Attention}~\cite{vaswani2017attention} and utilizes \emph{GroupNorm}~\cite{wu2018group}.

\noindent \textbf{Motion Generator} models the distribution of motion from the first frame to the random $n$-th frame, and it is implemented with another network $\rho_\phi(\cdot)$ with parameters $\phi$.
Therefore, the learning process minimizes the difference between $\brho_\phi(\bx(0)_0, n)$ and $\bx(n)_0$ as Figure~\ref{fig:pipeline} shows, which is similar to recent implicit neural function methods~\cite{yu2022generating,skorokhodov2021stylegan}.

By experimentally combining the two streams, one can model video data.
However, due to the complexity of video data, we observed that the basic implementation does not converge.  In addition, it losses significant generation quality and generates discontinuous motions.
Therefore, we propose to extend the aforementioned video generation process with the following improvements.

\subsection{Positional Group Normalization}
Our first key idea for improving the diffusion network is to incorporate the spatial and temporal positional encoding of 4D coordinates $(h,w,n,t)$ between each U-Net blocks, for modeling continuous changes in both the space $h,w$ and time $n$ with different diffusion timesteps $t$.
The correlation between spatial and temporal features crucially affects the continuity of video data but is conventionally ignored due to its complexity.
Empirically, such complexity can be decomposed for modeling in an iterative denoising process. We propose to directly incorporate the correlation into networks in a feature modulation manner, similar to AdaIN~\cite{karras2019style} and FiLM~\cite{perez2018film}.

The concept is illustrated in the right part of Figure~\ref{fig:pipeline}.
Specifically, the positional encoding mapped from 4D coordinates is extracted through an MLP (fully-connected neural network) with sinusoidal activation~\cite{sitzmann2020implicit} after its first layer.
Recent studies on implicit neural representations (INRs)~\cite{sitzmann2020implicit,tancik2020fourier} have shown that periodic activation is capable of modeling high dimensional space with coordinates.
Inspired by it, our introduced Positional Group Normalization (PosGN) based on group-norm~\cite{wu2018group} is defined as 
\begin{align}
  \alpha, \beta &= \text{MLP}(h,w,n,t)\\
  \text{PosGN}(x, \alpha, \beta) &= \alpha \cdot \text{GroupNorm}(x) + \beta,
\end{align}
where $x$ is the obtained feature from the U-Net blocks, $(\alpha, \beta)$ is a pair of affine transformation parameters extracted from the $\text{MLP}$, and it then scales and shifts feature $x$ using parameters $(\alpha, \beta)$.
PosGN is based on the empirical superiority of adaptive group normalization (AdaGN)~\cite{nichol2021improved}, which has been shown to benefit diffusion models, and the difference between them are the introduced periodic activated MLP and the additional spatial and temporal dimensions.
Compared with the recent INR-based work, PosGN is particularly suited for diffusion models.
It is because the noisy images are essential conditions that cannot be replaced by coordinates as INRs have done.
Besides, PosGN provides a hierarchical feature modulation when it is incorporated into the applied diffusion networks.

As a result, our proposed VIDM benefits from the capability of modeling spatial and temporal changes led by PosGN.
Based on the new paradigm, the learning objective of our motion generation extended from the content modeling~\eqref{eq:newobj} for an arbitrary $n$-th frame is formulated as 
\begin{align}
  \mathcal{L}_\phi(\bepsilon, \bx(n)_t, t, n) = \sqrt{(\brho_\phi(\bx(n)_t, t, n) - \bepsilon)^2 + \eta^2}.
\end{align}
Coordinates $(h,w)$ are derived from features on-the-fly and thus are not treated as the network input.
For convenience and efficiency, coordinates $(h,w,n,t)$ are only generated at the first time and then cached for the next running.
Therefore, PosGN shares a very similar computational cost as the vanilla AdaGN when the running times are large, which is natural to diffusion models.
In the rest of this paper, we use PosGN as our default settings and denote $\brho_\phi(\cdot, t, n)$ as $\brho_\phi(\cdot)$ for simplification.

\algrenewcommand\algorithmicindent{0.5em}%
\begin{figure}[t]
\begin{minipage}[t]{0.495\textwidth}
\begin{algorithm}[H]
  \caption{Motion Learning} \label{alg:training}
  \label{algo}
  \small
  \begin{algorithmic}[1]
    \State \textbf{input:} random frames $\{\bx(0), \bx(n-1), \bx(n)\}$ 
    \Repeat
      \State $t \sim \mathrm{Uniform}(\{1, \dotsc, T\})$
      \State $\bepsilon\sim\mathcal{N}(\bzero,\bI)$
      \State $\bz = \mathbf{v} (\bx(0)_0, \bx(n-1)_0)$
      \State $r = \hat\brho_\phi (\bx(0)_0)$
      \State $\bx(n)_t = \sqrt{\bar\alpha_t} \bx(n)_0 + \sqrt{1-\bar\alpha_t}\bepsilon$
      \State Take gradient descent step on
      \Statex $\qquad \grad_\phi \sqrt{(\bepsilon - \brho_\phi(\bx(n)_t, \bz) - r)^2 + \eta^2}$
    \Until{converged}
    \State \textbf{return:} motion network $\brho_\phi(\cdot)$
  \end{algorithmic}
\end{algorithm}
\end{minipage}
\hfill
\begin{minipage}[t]{0.495\textwidth}
\begin{algorithm}[H]
  \caption{Video Generation} \label{alg:sampling}
  \small
  \begin{algorithmic}[1]
    \For{$n=0, \dotsc, N-1$}
    \State $\bx(n)_T \sim \mathcal{N}(\bzero, \bI)$
    \For{$t=T, \dotsc, 1$}
      \State if $n = 0$
      \Statex $\qquad \bx(n)_{t-1} \sim \bepsilon_\theta(\bx(n)_t, c, t)$
      \State else
      \Statex $\qquad \bz = \mathbf{v} (\bx(0)_0, \bx(n-1)_0)$ 
      \Statex $\qquad r = \hat\brho_\phi (\bx(0)_0)$
      \Statex $\qquad \bx(n)_{t-1} \sim \brho_\phi(\bx(n)_t, \bz) + r$
    \EndFor
    \EndFor
    \vspace{.06in}
  \end{algorithmic}
\end{algorithm}
\end{minipage}
\vspace{-1\baselineskip}
\end{figure}

\subsection{Implicit Motion Condition}
Modeling long continuous video data has been a long-standing problem, even though we have seen the exploration in INRs and our proposed PosGN with positional encoding, the intermediate information between long video frames cannot be accurately represented.
Furthermore, from the results in the literature~\cite{yu2022generating,skorokhodov2021stylegan}, we find that the intermediate information plays a crucial role in the video continuation, otherwise, the generated long videos only contain nearly meaningless motions.

Our second idea is extended from the proposed PosGN, based on the time condition, instead of explicit coordinates, we propose to condition on the latent code of the latest frame and the first frame at the denoising process.
The latent code is an optical flow~\cite{horn1981determining} like feature estimated by an additional network $\mathbf{v}(\cdot)$, implemented as SpyNet~\cite{ranjan2017optical}, which has been demonstrated in motion extraction for video enhancement and interpolation.
To elaborate, a pretrained optical flow estimation network $\mathbf{v}(\cdot)$ is applied to estimate the latent $\mathbf{z}$ between frames $\mathbf{v}(\bx(0)_0, \bx(n-1)_0)$ for the $n$-th frame $\bx(n)_0$ generation, which is performed in an autoregressive manner.
Since the latent code is capable of ensembling the continuous motion feature that consistently exist in the denoising process, it can ensure that the intermediate information is implicitly incorporated into learning.
Therefore, the learning process with the implicit latent condition is 
\begin{align}
  \bz &= \mathbf{v}(\bx(0)_0, \bx(n-1)_0)\\
  \mathcal{L}_\phi(\bepsilon, \bx(n)_t, \bz) &= \sqrt{(\brho_{\phi}(\bx(n)_t, \bz) - \bepsilon)^2 + \eta^2}.
\end{align}
The parameters of $\mathbf{v}(\cdot)$ are updated with the diffusion networks together without specification.
The cost of conditioning on the latent at each denoising process is only increased at the first timesteps and can be then cached.
As will be shown in the ablation study, implicit learning is crucial for modeling long video data and can significantly improve the ultimate performance.

\subsection{Adaptive Feature Residual}
To further simplify the motion modeling complexity, we propose to model the residual of content features at each denoising timestep adaptively.
An additional encoder that shares the similar architecture of the diffusion network is utilized, and it conditions on the first frame $\bx(0)_0$ and timesteps $t$.
We denote the encoding as $\hat\rho_\phi(\cdot)$ and the residual feature as $r$, and thus network $\rho_\phi(\cdot)$ is actually learning to synthesize the residual, which significantly simplifies the learning at each timestep and enables better implicit motion learning.

The complete procedure of our method for both motion learning and video generation is detailed in Algorithm~\ref{algo}.
Remark that content generation learning is kept the same as DDPM except for the truncation trick and robustness penalty is applied for enhancing the generation capability.

\section{Experiments}

\textbf{Datasets and settings.}
Most datasets follow the protocols of their original papers except where specified.
To compare the visual quality of the results, we use the I3D network trained on Kinetics-400~\cite{kay2017kinetics} for reporting the Fr\'echet video distance (FVD)~\cite{unterthiner2018towards} performance, which measures the probability distribution difference between two groups of video results and is recognized by the other prior arts~\cite{yu2022generating, skorokhodov2021stylegan}.
For reference, we also report the Inception score (IS)~\cite{salimans2016improved} performance and Fr\'echet inception distance (FID)~\cite{heusel2017gans} following the evaluation procedure of DIGAN~\cite{yu2022generating}.
All evaluation is conducted on 2048 randomly selected real and generated videos for reducing variance. The experiments are conducted on \emph{UCF-101}~\cite{soomro2012ucf101}, \emph{Tai-Chi-HD}~\cite{siarohin2019first}, \emph{Sky Time-lapse}~\cite{xiong2018learning}, and \emph{CLEVRER}~\cite{yi2020clevrer}.

\paragraph{Baselines.}
The major baseline for comparison is DIGAN~\cite{yu2022generating}, which is the current state-of-the-art in video generation and is the first work that incorporates INRs.
We also compare the performance of our method with that of VGAN~\cite{vondrick2016generating}, TAGN~\cite{saito2017temporal}, MoCoGAN~\cite{tulyakov2018mocogan}, ProgressiveVGAN~\cite{acharya2018towards}, DVD-GAN~\cite{clark2019adversarial}, LDVD-GAN~\cite{kahembwe2020lower}, TGANv2~\cite{saito2020train}, MoCoGAN-HD~\cite{tian2021good}, VideoGPT~\cite{yan2021videogpt}, StyleGAN-V~\cite{skorokhodov2021stylegan}, VDM~\cite{ho2022video},  and TATS~\cite{ge2022long}.
We collect the performance score from the references or re-implemented results from DIGAN and StyleGAN-V if available.
For the CLEVRER performance, we train DIGAN and StyleGAN-V with their official code and our implementation with the same settings.

\begin{figure}[htbp]
  \centerline{\includegraphics[width=\linewidth]{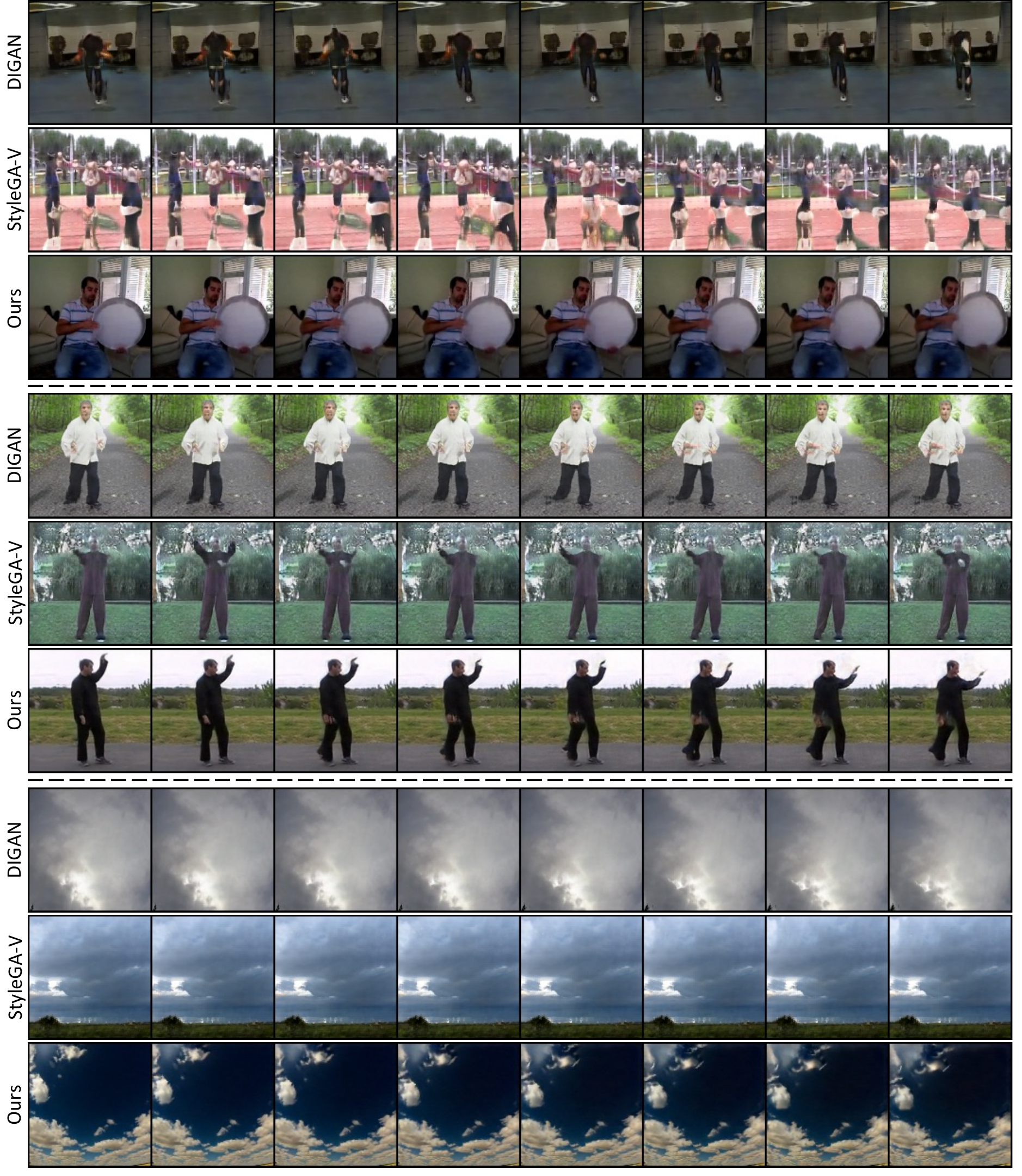}}
  \vspace{-.5\baselineskip}
  \caption{Sample result comparisons on the  \emph{256-UCF101$^{16}$}, \emph{128-TaiChi$^{16}$}, and \emph{256-SkyTimelapse$^{16}$} datasets. Each presented frame is selected with 2 frames interval.}
  \label{fig:mainr}
      \vspace{-1em}
\end{figure}

\begin{table*}[htbp]
  \centering
  \resizebox{.88\linewidth}{!}{
  \begin{tabular}{lcccccccc}
  \toprule
  & MoCoGAN$^\dagger$ & MoCoGAN-HD & VideoGPT & DIGAN & DIGAN$^\ddagger$ & StyleGAN-V & TATS & VIDM \\
  & \emph{CVPR18} & \emph{ICLR21} & \emph{arXiV21} & \emph{ICLR22} & \emph{ICLR22} & \emph{CVPR22} & \emph{ECCV22} & (ours) \\
  \midrule
  ~\textit{256-UCF101}$^{16}$~ & 1821.4 & 1729.6 & 2880.6 & 1630.2 & 471.9 & 1431.0 & 332 & \textbf{294.7} \\
  \textit{256-UCF101}$^{128}$ & 2311.3 & 2606.5 & \emph{N/A} & 2293.7 & \emph{N/A} & 1773.4 & - & \textbf{1531.9} \\
  \textit{256-SkyTimelapse}$^{16}$ & 85.9 & 164.1 & 222.7 & 83.1 & 83.1 & 79.5 & 132 & \textbf{57.4} \\
  \textit{256-SkyTimelapse}$^{128}$ & 272.8 & 878.1 & \emph{N/A} & 196.7 & 196.7 & 197.0 & - & \textbf{140.9} \\
  \bottomrule
  \end{tabular}}

  \begin{subtable}{.49\textwidth}
  \raggedleft
  \resizebox{.89\linewidth}{!}{\begin{tabular}[t]{lccc}
  \toprule
  & DIGAN & StyleGAN-V & VIDM (ours) \\ 
  \midrule
  \textit{256-CLEVRER}$^{16}$ & 112.5 & 106.1 & \textbf{87.4} \\
  \textit{256-CLEVRER}$^{128}$ & 531.7 & 493.3 & \textbf{426.5}  \\
  \bottomrule
  \end{tabular}
  }
  \end{subtable}
  \begin{subtable}{.48\textwidth}
    \raggedright
    \resizebox{.89\linewidth}{!}{\begin{tabular}[t]{lccc}
    \toprule
    & DIGAN & StyleGAN-V & VIDM (ours) \\ 
    \midrule
    \textit{128-TaiChi}$^{16}$ & 128.1 & 143.5 & \textbf{121.9}  \\
    \textit{128-TaiChi}$^{128}$ & 748.0 & 691.1 & \textbf{563.6}  \\
    \bottomrule
    \end{tabular}
    }
  \end{subtable}
  \vspace{-.5\baselineskip}
  \caption{Fr\'echet video distance~\cite{unterthiner2018towards} comparison. The compared methods are re-trained on the CLEVRER dataset by us, and by \citet{skorokhodov2021stylegan} and \citet{yu2022generating} on the other datasets with their official implementation. MoCoGAN$^\dagger$ is implemented with StyleGAN2 as its backbone. DIGAN$^\ddagger$ is class conditional.}
  \label{table:main-results}
\end{table*}

\begin{table*}[htbp]
  \centering
  \resizebox{.9\linewidth}{!}{\begin{tabular}{lcccccccc}
  \toprule
  & \multicolumn{8}{c}{\cellcolor{gray! 20} \textit{Train split}} \\
  & VGAN & TGAN & MoCoGAN & ProgressiveVGAN & LDVD-GAN & VideoGPT & TGANv2 & DIGAN \\
  & \emph{NeurIPS16} & \emph{ICCV17} & \emph{CVPR18} & \emph{arXiv18} & \emph{NN20} & \emph{arXiv21} & {IJCV20} & \emph{ICLR22} \\
  \midrule
  \textit{128-UCF101$^{16}$ IS ($\uparrow$)} & 8.31\stdv{.09} & 11.85\stdv{.07} & 12.42\stdv{.07} & 14.56\stdv{.05} & 22.91\stdv{.19} & 24.69\stdv{.30} & 28.87\stdv{.67} & 29.71\stdv{.53} \\
  \textit{128-UCF101$^{16}$ FID ($\downarrow$)} & - & - & - & - & - & - & 1209\stdv{28} & 655\stdv{22} \\
  \bottomrule
  \end{tabular}}
  \vspace{0.05in}

  \resizebox{.9\linewidth}{!}{\begin{tabular}{lccccccccc}
  \toprule
  & \multicolumn{2}{c}{\cellcolor{gray! 20} \textit{Train split}} & \multicolumn{7}{c}{\cellcolor{gray! 40} \textit{Train+test split}} \\
   & VIDM & VIDM$^\dagger$ & DVD-GAN & MoCoGAN-HD & DIGAN & StyleGAN-V & DIGAN$^\ddagger$ & VDM & VIDM$^\dagger$ \\
   & \emph{Ours} & \emph{Ours} & \emph{arXiV19} & \emph{ICLR21} & \emph{ICLR22} & \emph{CVPR22} & \emph{ICLR22} & \emph{arXiv22} & Ours \\
  \midrule
  \textit{128-UCF101$^{16}$ IS ($\uparrow$)} & \emph{53.34} & 35.20 & 27.38\stdv{.53} & 32.36 & 32.70\stdv{.35} & 32.70\stdv{.35} & \emph{59.68\stdv{.45}} & 57\stdv{.62} & \textbf{64.17} \\
  \textit{128-UCF101$^{16}$ FID ($\downarrow$)} & 306 & \emph{471} & - & 838 & 577\stdv{21} & - & - & \emph{295\stdv{3}} & \textbf{263} \\
  \bottomrule
  \end{tabular}}
  \vspace{-.5\baselineskip}
  \caption{IS and FVD comparisons. For fair comparisons, we re-train our VIDM without video class condition, named VIDM$^\dagger$.}
  \label{tab:isfid}
\vspace{-1\baselineskip}
\end{table*}

\paragraph{Diffusion Network.}
The diffusion network architecture of our method is an autoencoder network that follows the design of PixelCNN++~\cite{salimans2017pixelcnn++}.
We apply multiple multi-head attention modules~\cite{vaswani2017attention} at features in a resolution of $16\times 16$ for capturing long-range dependence that benefits the perceptual quality.
It has been verified by DDPM~\cite{ho2020denoising} and its variants~\cite{dhariwal2021diffusion, nichol2021improved}, and we keep minimal changes.

\paragraph{Main Results.}
We present the main quantitative results comparison in Table~\ref{table:main-results} and Table~\ref{tab:isfid}, and the main qualitative results comparison is Figure~\ref{fig:mainr}. We remark that our performance significantly outperforms the very recent state-of-the-art DIGAN and StyleGAN-V in all of the video data as can be seen from the two tables. Among them, \emph{128-TaiChi} and \emph{256-UCF101} is the hardest video data since their movement is minimal and Frames Per Second (FPS) is varying between videos, but our method can still achieve comparable performance and even better without discriminators.

\paragraph{Ablations.}
Multiple potential design choices are available in our final method, and most of them affect the results to some degree.
We ablate the core components and show the details in Table~\ref{table:abc} for content generator ablations and Table~\ref{table:abm} for motion generator ablations.
As the results are shown in Table~\ref{table:abc}, the removed sampling space truncation and robustness penalty hurt the performance of content modeling.
These results also verify that removing the robustness penalty decreases both the content modeling ability and motion modeling ability.

\begin{figure*}[htbp]
    \centering
    \includegraphics[width=\linewidth]{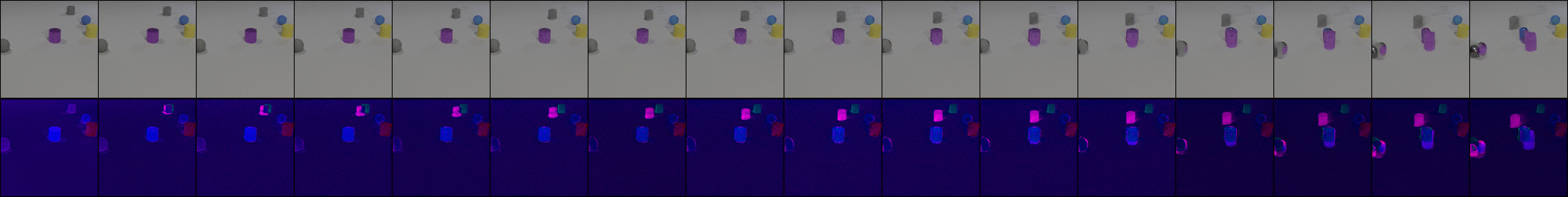}
    \vspace{-1.5\baselineskip}
    \caption{Visualization of the generated latent and its corresponding frames.}
    \label{fig:latent}
    \vspace{-1em}
\end{figure*}

For the motion generator, we measure the ablation effects by comparing generated videos in a varying number of frames, which is the most representative score for measuring continuous and smoothness differences.
In Table~\ref{table:abm}, we remove the positional group normalization and implicit motion conditions to see the difference. It is surprising that the modeling capability severely depends on the two proposed components, especially for long video generation.
From the results visualized in Figure~\ref{fig:ab}, we can notice that simply applying diffusion models (i.e., \emph{Ablation1}) without modification can only generate static images. Applying implicit conditions without PosGN (i.e., \emph{Ablation2}) faces the same issue since they cannot model the spatial and temporal changes. In contrast, even though applying PosGN without implicit conditions (i.e., \emph{Ablation3}) can help the network generates different frames, its results are still noncontinuous.
In Figure~\ref{fig:latent}, we visualize the latent and its corresponding video frames for further clarification.

\begin{table*}[htbp]
  \centering\small
  \begin{subtable}{.42\textwidth}
  \centering\small
  \begin{tabular}[t]{lccc}
  \toprule
   & FID & IS & FVD$^{16}$ \\ 
  \midrule
  vanilla one & 23.0 & 3.04 & 115.4  \\
  w/o \emph{sampling space truncation} & 21.1 & 3.07 & 107.9  \\
  w/o \emph{robustness penalty} & 19.4 & 3.07 & 95.5 \\
  \midrule
  default VIDM & 18.4 & 3.07 & 87.4 \\
  \bottomrule
  \end{tabular}
  \caption{Ablation study regarding content generator.}
  \label{table:abc}

  \begin{tabular}[t]{lccc}
  \toprule
  & FVD$^{16}$ & FVD$^{64}$ &  FVD$^{128}$ \\ 
  \midrule
  vanilla one & 603.7 & 610.0 & 648.7  \\
  w/o \emph{PosGN} & 532.1 & 581.3 & 604.5  \\
  w/o \emph{Implicit Conditions} & 584.8 & 552.1  & 614.1  \\
  \midrule
  default VIDM & 87.4 & 286.6 & 426.5 \\
  \bottomrule
  \end{tabular}
  \caption{Ablation study regarding motion generator.}
  \label{table:abm}
  \end{subtable}
  \begin{subtable}{.42\textwidth}
  \centering\small
  \includegraphics[width=\textwidth]{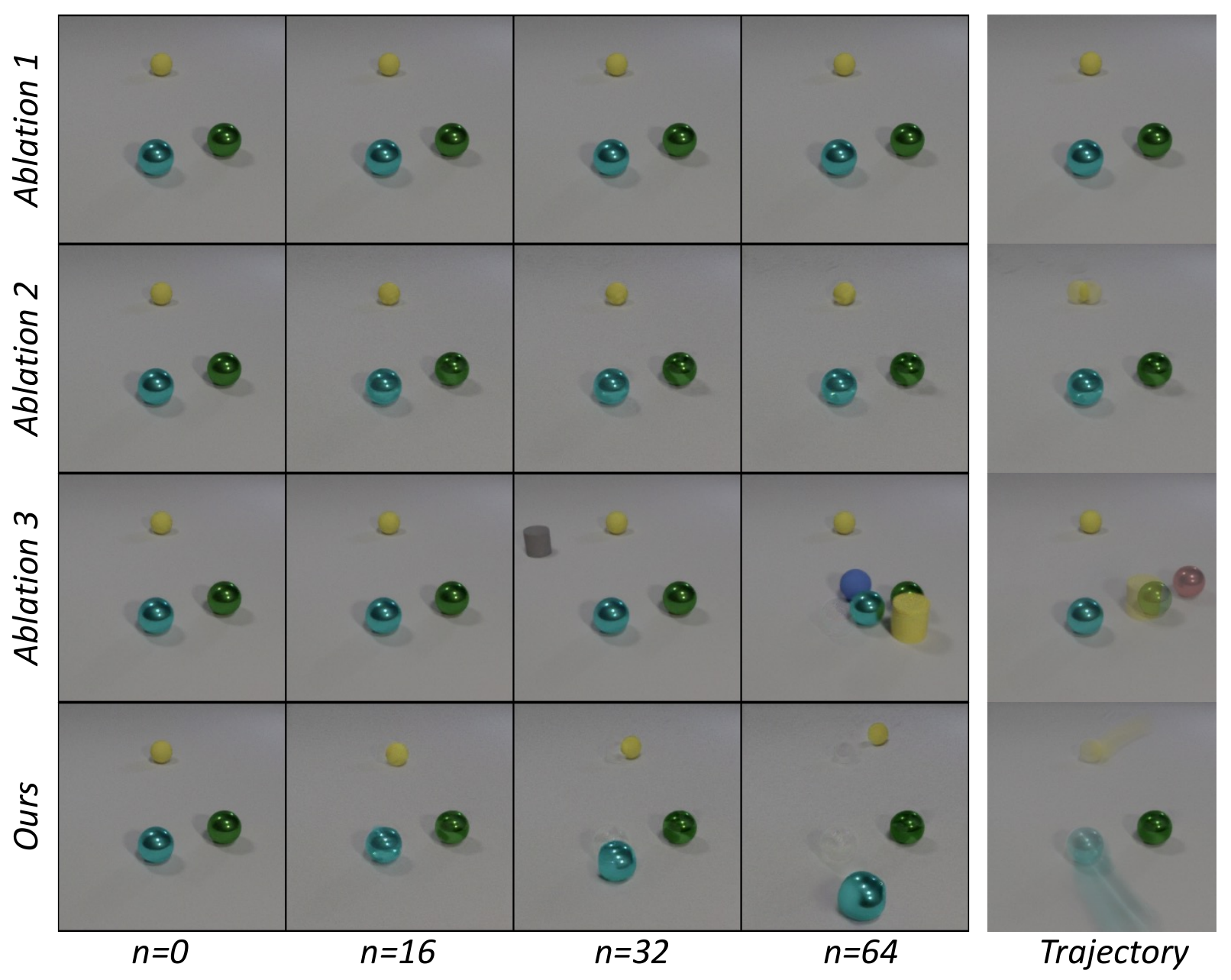}
  \vspace{-1\baselineskip}
  \caption{Ablation results in different settings.}
  \label{fig:ab}
  \end{subtable}
  \vspace{-1em}
\caption{Ablations on different settings with quantitative and qualitative results.}
\vspace{-1\baselineskip}
\end{table*}

\section{Related Work}
\paragraph{Generative Models.}
Existing generative models can be categorized into likelihood-based and implicit models, based on the way of representing probability distribution.
Among them, Variational Auto-encoders (VAEs)~\cite{kingma2013auto}, Autoregressive models~\cite{van2016pixel, germain2015made}, Normalizing Flow~\cite{dinh2016density}, and Diffusion models~\cite{sohl2015deep, ho2020denoising} directly model the probability distribution of data via maximum likelihood.
In contrast, GANs~\cite{goodfellow2014generative} implicitly represent the probability distribution via their sampled results.
Though the idea of GANs is simple, the boundary has been significantly pushed by GANs and their representative variants, including StyleGAN~\cite{karras2019style, karras2020analyzing, karras2020training} and BigGAN~\cite{brock2018large}.
Moreover, many general techniques based on GANs have emerged, including $R_1$ regularization~\cite{mescheder2018training}, path length regularization~\cite{karras2020analyzing}, truncation trick~\cite{karras2019style}, spectral normalization~\cite{miyato2018spectral}, image inversion~\cite{mei2021ltt}, and adaptive discriminator~\cite{karras2020training}.
However, we find that such techniques are rarely explored in diffusion models.
In this paper, inspired by these heavily refined techniques, we introduce several adaptive methods for refining diffusion models and observe that they benefit for both video and image generation.

\paragraph{Conditional Generative Models.}
Modeling the probability distribution of complex datasets such as ImageNet~\cite{deng2009imagenet} can face potential training instability and mode collapse issues.
Therefore, the way of leveraging additional conditions as a guidance is explored and becoming the most promising way of mitigating the issues.
For GANs, class information can be fed into the generator~\cite{mirza2014conditional, odena2017conditional, de2017modulating, dumoulin2016learned, brock2018large} and the discriminator~\cite{miyato2018cgans,karras2019style} for fascinating class-conditional sampling.
For diffusion models, the class condition shows a better performance boost as the class embeddings used in DDPM~\cite{ho2020denoising}.
Furthermore, resulting from the iterative denoising process of diffusion models, which enables hierarchical conditional features, utilizing the class feature of noisy images of different time steps can help diffusion models achieve the new art~\cite{dhariwal2021diffusion}.
Different from class conditions, the modality of conditions could be images~\cite{ledig2017photo, nair2022ddpm} and even texts like DALLE~\cite{ramesh2021zero} for different aims.
VIDM is the first work that explores the implicit conditions for video generation, and it also benefits from the iterative denoising process of diffusion models, which allows hierarchical conditional features of complex spatial-temporal changing.

\paragraph{Video Generation.}
Video generation has been dominated by 3D CNNs~\cite{tran2015learning} for a long time until the recent emergence of Implicit Neural Representation (INR)~\cite{sitzmann2020implicit, tancik2020fourier}.
The early 3D CNNs based video generation works take all frames of the video as a single point on the video subspace.
They then generate a cuboid as the result of each sampling process~\cite{vondrick2016generating, saito2017temporal}, and such a manner has been extended into the recent diffusion fashion~\cite{ho2022video, harvey2022flexible}.
However, this line can hardly achieve desired results due to the difficulty of modeling spatial-temporal changing, and their scalability is significantly limited according to the cubic complexity~\cite{saito2020train}.
Later work decomposes the generation process into content and motion separately~\cite{tulyakov2018mocogan, clark2019adversarial, tian2021a, fox2021stylevideogan}, which simplifies the learning but still requires the discriminator to apply 3D CNNs on extracting temporal features.
The other line of video generation~\cite{yu2022generating,skorokhodov2021stylegan} based on INRs is similar to the image generation applications work~\cite{skorokhodov2021adversarial}, which adds additional temporal dimension at the coordinates and thus can process each frame separately.
However, such an INR protocol can hardly be applied to diffusion models.
Therefore, our method incorporates the coordinate embeddings of INRs as the normalization and conditions on the implicit latent.
We experimentally find that the new paradigm benefits the continuous of the generated complex videos.

\paragraph{Limitations and Ethics Statement.}
The major limitation of this work comes from the efficiency issue of diffusion models.
Limited by the expressibility of the Gaussian process, multiple iterative denoising process is required before producing plausible results. 
Therefore, the complexity of video generation consists of the number of video frames and the number of diffusion time steps.
The potential negative societal impacts of this work come from the generated unethical videos.
These generated videos a.k.a Deepfake have emerged as an important social issue and attracted great attention.
However, we are happy to see that significant funds and efforts have been devoted to detecting these fake videos, including DARPA's Semantic Forensics program which is highly inspired by the StyleGAN series~\cite{karras2019style}.
Our work can be useful in promoting them.


\section{Conclusion}
In this work, we proposed a new diffusion probabilistic model for video data, which provides a unique implicit condition paradigm for modeling continuous spatial-temporal changing of videos.
The model is capable of sampling frames according to latent that encodes dynamics.
Comprehensive experiments on the high-resolution, long video data demonstrated our method not only with visual quality superiority but also better diversity.
We hope the work would benefit and inspire both video generation and conditional diffusion models as a strong baseline in the future. 

\section{Acknowledgement}
Research was sponsored by the Army Research Laboratory and was accomplished under Cooperative Agreement Number W911NF-21-2-0211. The views and conclusions contained in this document are those of the authors and should not be interpreted as representing the official policies, either expressed or implied, of the Army Research Office or the U.S. Government. The U.S. Government is authorized to reproduce and distribute reprints for Government purposes notwithstanding any copyright notation herein.
We thank Meihan Wei for helpful feedbacks.

\setcitestyle{numbers}
\bibliography{main}

\end{document}